\documentclass[conference]{IEEEtran}
\IEEEoverridecommandlockouts
\usepackage{amsmath,amssymb,amsfonts}
\usepackage{algorithmic}
\usepackage{graphicx}
\usepackage{textcomp}
\usepackage{xcolor}
\usepackage[noadjust]{cite}
\usepackage{multirow}
\usepackage{booktabs} 
\usepackage{threeparttable}
\graphicspath{{figures/}} 
\def\BibTeX{{\rm B\kern-.05em{\sc i\kern-.025em b}\kern-.08em
    T\kern-.1667em\lower.7ex\hbox{E}\kern-.125emX}}
\usepackage[,
            linkcolor=blue,       
            anchorcolor=blue, 
            citecolor=blue,
            ]{hyperref}
\begin{document}

\title{FeatureNorm: L2 Feature Normalization for Dynamic Graph Embedding
}

\author{
\IEEEauthorblockN{Menglin Yang, Ziqiao Meng, and Irwin King}
Department of Computer Science and Engineering\\
The Chinese University of Hong Kong, Shatin, N.T., Hong Kong\\
\{mlyang, zqmeng, king\}@cse.cuhk.edu.hk \\
}
\maketitle
\begin{abstract}
Dynamic graphs arise in a plethora of practical scenarios such as social networks, communication networks, and financial transaction networks. Given a dynamic graph, it is fundamental and essential to learn a graph representation that is expected not only to preserve structural proximity but also jointly capture the time-evolving patterns.  Recently, graph convolutional network (GCN) has been widely explored and used in non-Euclidean application domains. The main success of GCN, especially in handling dependencies and passing messages within nodes, lies in its approximation to Laplacian smoothing. As a matter of fact, this smoothing technique can not only encourage must-link node pairs to get closer but also push cannot-link pairs to shrink together, which potentially cause serious feature shrink or oversmoothing problem, especially when stacking graph convolution in multiple layers or steps. For learning time-evolving patterns, a natural solution is to preserve historical state and combine it with the current interactions to obtain the most recent representation. Then the serious feature shrink or oversmoothing problem could happen when stacking graph convolution explicitly or implicitly according to current prevalent methods, which would make nodes too similar to distinguish each other. To solve this problem in dynamic graph embedding, we analyze the shrinking properties in the node embedding space at first, and then design a simple yet versatile method, which exploits L2 feature normalization constraint to rescale all nodes to hypersphere of a unit ball so that nodes would not shrink together, and yet similar nodes can still get closer. Extensive experiments on four real-world dynamic graph datasets compared with competitive baseline models demonstrate the effectiveness of the proposed method.
\end{abstract}

\begin{IEEEkeywords}
Dynamic graph embedding, Feature shrink, Normalization, Graph convolutional network
\end{IEEEkeywords}

\section{Introduction}
Graphs are ubiquitous non-Euclidean data structures applied in various scenes, such as social networks, communication networks, and financial transaction networks, etc. Learning node representations in a latent space while preserving structural properties and interactive information stored in the original graph is one of the fundamental problems, which has attracted much attention in machine learning communities. 

Most existing node embedding works assume the graph is static and is associated with a fixed set of nodes and edges~\cite{ge:deepwalk},~\cite{ge:line},~\cite{ge:node2vec}. However, networks in many real-world application scenarios are intrinsically time-evolving. Evidence can be found in social networks, (traffic) communication networks, and financial transaction networks. Modeling dynamic graphs is challenging due to complicated cases brought by an evolving process: nodes can be added or removed, edges can appear or disappear, and communities can be merged or split. For instance, in social and email communication networks, people tend to frequently add and sometimes remove their connections based on daily business affairs; in IP-IP networks, agents periodically send messages from one address to any other address in the whole network, which makes the temporal model tough to capture the true evolving regularities. In short, the inherent difficulties stated above result in a lack of convincing dynamic graph models so far.

A surge of graph research works emerged after the simplified graph convolutional neural network (GCN) was proposed by Kipf and Welling~\cite{gcn:kipfGCN}. However, the latest developments on GCN show that applying multiple graph convolutional layers would introduce oversmoothing problem, i.e., nodes in a graph become more and more similar thus being indistinguishable. The reason is that graph convolution is a special type of Laplacian smoothing. By stacking graph convolutional layers repeatedly, the feature space will shrink together and the final node embedding matrix will converge to a low-rank matrix, which results in nodes being almost identical and indistinguishable representation. 
It was first introduced by Li, et al.~\cite{gcn_opt:deepanalysis} and further studied by~\cite{gcn_opt:pairnorm},~\cite{gcn_opt:jknet},~\cite{gcn_opt:pagerank}. 
Yet, we notice that very few works investigate this problem in dynamic graph models, although it harasses the development of dynamic graph research. In fact, in dynamic graph setting, stacking multiple graph convolutional layers to capture topological and temporal properties is an indispensable procedure according to present widely-used methods, but serious feature shrink or oversmoothing problems will also occur. Here we use the term ``serious feature shrink'' to express the same meaning as oversmoothing phenomenon, but at the same time to emphasize the cause of oversmoothing.

In short, in this paper we first analyze the feature shrink and oversmoothing problems in the dynamic graph embedding. Then we propose a simple yet versatile method, that is L2 feature normalization, to tackle the problem. Finally, extensive experiments show that the performance can be remarkably improved by the proposal. 
To summarize, the main contributions of our work are as follows:

\begin{itemize}
  \item We analyze the feature shrink and oversmoothing problems in dynamic graph embedding, which is the first work to the best of our knowledge.
  \item An effective and versatile feature normalization method is proposed to tackle serious feature shrink or oversmoothing issue in dynamic graph embedding.
  \item Comprehensive experiments demonstrate that our proposed method can not only improve the performance of various baselines obviously, but also can be served as a plug-in to boost the related models and unleash their true capabilities.
\end{itemize}

\section{Related works}


Dynamic graph embedding is an extension of static node embedding with an additional attention on the temporal-evolving information. Related works are generally carried out from two aspects: topological dependencies and temporal-evolving patterns.\\

\noindent
\textbf{Topological dependencies learning}. Early works for node embedding mainly center on factorization-based approaches, resorting to dimensionality reduction of graph Laplacian matrix~\cite{ge:spetral1},~\cite{ge:spetral2},~\cite{ge:spetral3},~\cite{ge:DANE}. These methods suffer from time and memory efficiency. To improve scalability, the random walk-based methods transfer the embedding into a network mining task, which is inspired by the success in the natural language process~\cite{tgcn:netwalk}. 
Recently, the generalizations of convolutions over graph-like data have achieved remarkable success. Defferrard et al.~\cite{gcn:cheb} first define graph convolutions using Chebyshev polynomial and reduce the computation significantly.
Kipf and Welling~\cite{gcn:kipfGCN} simplify graph convolution using the first-order Chebyshev polynomial filters with an symmetric normalized adjacency matrix. Hamilton et al.~\cite{gcn:graphsage} propose GraphSAGE for inductive learning.
Peter et al.~\cite{gcn:gat} further put forward a graph attention network (GAT) to explore the multiple attention mechanism in neighborhood message passing. Although these methods are proposed in a static graph, they are currently widely used in the structure learning of dynamic graph embedding~\cite{tgcn:dysat}, \cite{tgcn:evolvegcn},~\cite{tgcn:gated_gcn}, \cite{tgcn:tgat},~\cite{tgcn:traffic_gcn}, \cite{zhang2018gaan}.\\

\noindent
\textbf{Temporal-evolving patterns learning}. To learn the temporal-varying patterns, most works adopt an temporal-smoothness or recurrent learning fashion (definitions see \ref{section:definition of smoothness}), which is in line with the nature of the dynamism. In related literature, BCDG~\cite{tgcn:scalable} utilizes matrix decomposition approach, which explores the smoothness across consecutive time steps to model the time-evolving patterns.
NetWalk~\cite{tgcn:netwalk} takes a random walk method, using an autoencoder framework and minimizing the pairwise distance of vertex representation of each walk.
DynGEM~\cite{tgcn:dyngem} also adopts the autoencoder framework that constrains the local and global structure proximity.
With the success of graph convolution, most recent works mainly leverage an integrated GCN model to capture the structure information. The main difference lies in the temporal regularities learning.
DySAT~\cite{tgcn:dysat} brings self-attention mechanisms in both temporal and structural aspects. 
EvolveGCN~\cite{tgcn:evolvegcn} models the dynamism of the dynamic graph by using an RNN to capture the GCN parameters.\\

\noindent
\textbf{Oversmoothing in GNNs}.
Our work is also related to oversmoothing problem, and related works are as follows.
Li et al.~\cite{gcn_opt:deepanalysis} first pay attention to the oversmoothing problem, and they prove that graph convolution is a special case of Laplacian smoothing. In other words, applying graph convolutional layers repeatedly will result in the embedding feature being indistinguishable. Xu et al.~\cite{gcn_opt:jknet} propose a jumping knowledge network by utilizing skip connections for multi-hop message passing to overcome oversmoothing.
Klicpera et al.~\cite{gcn_opt:pagerank} put forward a customized Pagerank to propagate messages. 
Klicpera et al.~\cite{gcn_opt:resnet} consider adding residual connections like ResNet~\cite{resnet}, using skip scheme to connect the cross-layer features.
Rong et al.~\cite{gcn_opt:dropedge} argue that randomly removing a certain fraction of edges can reduce the oversmoothing problem. 
Zhao and Akoglu~\cite{gcn_opt:pairnorm} use the total pairwise distance to avoid the feature being washed away. 
The above methods mainly design for static graphs and do not consider the temporal condition. Our work is to tackle oversmoothing, more precisely i.e., serious feature shrink in dynamic graph settings with temporal information.

\section{Proposed Method}
\begin{figure}[t!]
  \includegraphics[width=\linewidth]{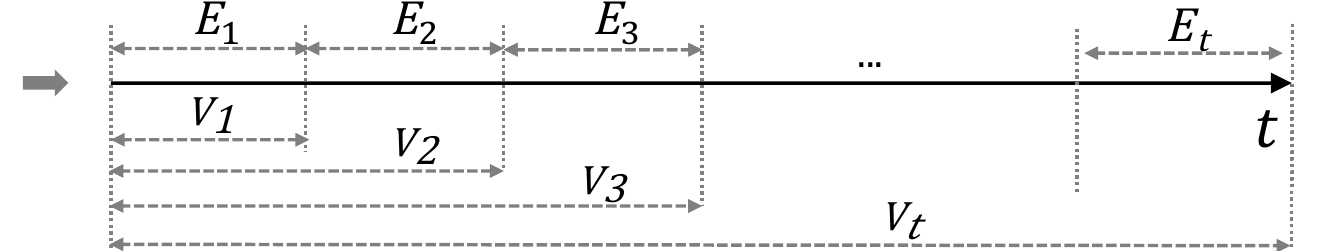}
  \caption{Slicing scheme of nodes and edges in dynamic graph, where the edges reflect current connections and topology, and node set contains ever appeared nodes till present time step.}
  \label{fig:time slicing_scheme}
\end{figure}
\subsection{Problem Definition}
\label{section:definition of smoothness}
Given a dynamic graph ${G}$ with $T$ graph snapshots $\{G_1, \cdots, G_t, \cdots, G_T\} $. Each snapshot $G_t = (V_t, E_t)$ is a weighted or unweighted, directed or undirected graph. $V_t$ is the observed node set from beginning to the time step $t$, meaning that we can handle variable-sized input, which is more reasonable for realistic scenes because the future new nodes are always agnostic to the current time step $t$. The number of nodes is non-decreasing since we need to learn their representations for further prediction in case that they appear again. $E_t$ denotes the links that appear in the current time interval [$t-1$, $t$], where the slicing scheme is illustrated in Figure~\ref{fig:time slicing_scheme}. The goal of dynamic graph embedding is to acquire the node representation at time step $t$ that can not only preserve topological dependencies but also capture the time-varying behaviors. 

\begin{figure}[!tp]
  \includegraphics[width=\linewidth]{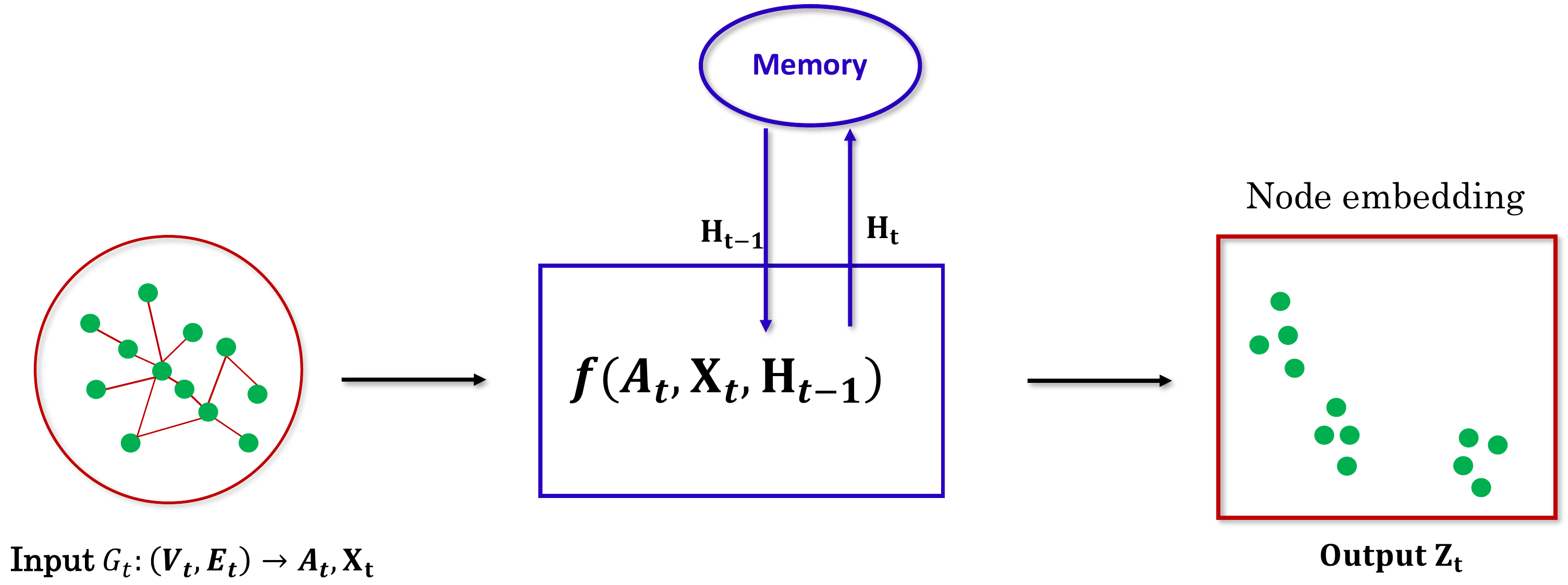}
  \caption{Basic learning paradigm for dynamic graph embedding. From current $V_t$ and $E_t$, we can extract the adjacency matrix $\mathbf{A}_t$ and then feed it into non-linear function $f$ together with node feature $\mathbf{X}_t$. To capture long-distance dependency, a memory cell is added to utilize historical state $\mathbf{H}_{t-1}$ and preserve current hidden state $\mathbf{H}_t$. The output node embedding $\mathbf{Z}_t$ is expected to jointly acquire topological information and temporal-evolving patterns.}
  \label{fig:paradigm_dge}
\end{figure}

Dynamic graph embedding can be summarized as the paradigm illustrated in Figure~\ref{fig:paradigm_dge}. At each time step $t$, the input is the adjacency matrix $\mathbf{A}_t\in\mathbb{R}^{n\times n}$ constructed from $E_t$ which reflects the connections within nodes and determines the topology of $G_t$, and the corresponding node features $\mathbf{X}_t\in \mathbb{R}^{n\times d_{i}}$, where $n$ is the number of nodes and $d_i$ is dimension of the \underline{i}nput feature. Dynamic graph modeling can be viewed as learning a mapping function $f$ given the current network $(\mathbf{A}_t, \mathbf{X}_t)$ and historical hidden state $\mathbf{H}_{t-1}\in \mathbb{R}^{n \times d_{h}}$ to learn a low-dimension representation $\mathbf{Z}_t \in \mathbb{R}^{n \times d_{o}}$, where $d_{h}$ and $d_{o}$ is the dimension of \underline{h}idden state $\mathbf{H}_{t-1}$, and \underline{o}utput node embedding $\mathbf{Z}_t$, respectively. The learning paradigm can be formalized as: 
\begin{equation}
\label{DyNGCN: equ:}
\mathbf{Z}_t, \mathbf{H}_t=f(\mathbf{A}_t, \mathbf{X}_t, \mathbf{H}_{t-1}). 
\end{equation}

Here, $f$ can be any nonlinear function. According to different temporal regularizers, we divide current prevent methods into two fashions: recurrent learning and temporal-smoothness learning. In recurrent learning, $f$ is instantiated as a recurrent network, e.g., RNN, GRU and LSTM \cite{Rnn, LSTM}, where the transformations inside it are replaced by graph convolutions. The output node embedding $\mathbf{Z}_t$ is expected to jointly acquire topological information and temporal-evolving patterns. Beyond that, considering the stability and continuity of the node embedding in two consecutive times, some works~\cite{tgcn:gated_gcn},~\cite{tgcn:icdm_traffic},~\cite{tgcn:dyngem} either (1) use historical state $\mathbf{H}_{t-1}$ to initialize $\mathbf{X}_t$ directly; (2) add temporal regularizer~\cite{tgcn:scalable} additionally; or (3) make alignment between two consecutive time steps~\cite{tgcn:tnodeembedding}, which are termed as temporal-smoothness learning in this paper. Temporal-smoothness learning can enhance nodes inter-dependency, and also facilitate the acquisition of robust and stable embeddings. These two learning fashions are typical methods in the current research field.

\subsection{Graph Convolutional Layer}
In this part, we mainly introduce how to learn the structural representation of the graph in snapshot $t$, so we temporarily ignore the subscript $t$ for clarity. At time step $t$, given a snapshot $G = (V, E)$ and we denote $\mathbf{A}$, $\mathbf{D}$ as the adjacency matrix and degree matrix, respectively. By applying a two-layer GCN \cite{gcn:kipfGCN, gcn:vae}, we obtain the output structural node embeddings:
\begin{equation}
\mathbf{Z} = \mathbf{\widetilde{A}}\mathrm{ReLU}(\mathbf{\widetilde{A}}\mathbf{X} \mathbf{W}_1)\mathbf{W}_2,
\end{equation}
where $\mathbf{\widetilde{A}}= \mathbf{\widetilde{D}}^{-\frac{1}{2}}\mathbf{\widehat{A}}\mathbf{\widetilde{D}}^{-\frac{1}{2}}$ is a symmetrically normalized adjacency matrix, $\mathbf{\widehat{A}}=\mathbf{A} + \mathbf{I}$ and $\mathbf{\widetilde{D}}=\mathbf{D}+\mathbf{I}$ are the augmented adjacency and degree matrices with added self-loops, respectively. $\mathbf{W}_1$ and $\mathbf{W}_2$ are the learnable matrices in GCN. Graph convolution can be interpreted as message fusion within connected nodes. 

In the analysis of GCN's working mechanism, Li et al.~~\cite{gcn_opt:deepanalysis} show that a graph convolution is a special form of Laplacian smoothing. This smoothing technique can force the representation of similar node pairs to be more tight. Therefore, GCN is able to obtain a pleasant performance in many tasks. Though, we also need to be aware that the above smoothing technique can push cannot-link node pairs to get together, especially when stacking graph convolution step by step or layer by layer. In this case, the node representation would become too smooth to be distinguished with each other, leading oversmoothing phenomenon. 

By decomposing GCN, we can find that the smoothing mainly lies in feature shrinking properties. In dynamic graph embedding, to obtain robust and impressive performance, stacking multiple layers implicitly or explicitly is inevitable which would increase feature shrink gradually and potentially causes oversmoothing problem. In the following section, we will start with analyzing feature shrink.

\subsection{Feature Shrink and Oversmoothing}
\label{section: oversmoothing}
We begin with easily splitting one-step graph convolution operation into two basic steps: (1) aggregation step; (2) transformation step. The message passing in GCN~\cite{gcn:kipfGCN} for a center node $v_i$ is formulated as:
\begin{equation}
\mathbf{z}_i=\sigma\left(\sum_{v_j\in N(v_i)}\tilde{a}_{ij}\mathbf{x}_jW\right),
\end{equation} 
where $\tilde{a}_{ij}=a_{ij}/d_{ii}$ is a normalized weight between node $v_j$ and $v_i$, $d_{ii}$ is the degree of node $v_i$. This rule can be decomposed into two steps:\\
(1) Aggregation step. Aggregating message from the neighbors, $N(v_i)$ of node $v_i$, where we treat each node is the neighbor of itself as well, and thus we have:
\begin{equation}
\mathbf{\tilde{x}}_i=\sum_{v_j \in N(v_i)}\tilde{a}_{ij}\mathbf{x}_j.
\end{equation}
(2) Transformation step. Transforming the aggregated state $\mathbf{\tilde{x}}_i$ into a new space with a leanerable matrix $W$ and nonlinear activation function,
\begin{equation}
\mathbf{\tilde{z}}_i=\sigma(\mathbf{\tilde{x}}_iW).
\end{equation}
Then it can be shown that the overall distance of node embedding is reduced in aggregation step, see Theorem 1.\\

\noindent  
\textbf{Theorem 1}~\cite{labelpropagation}
\textit{Let the overall distance of node embeddings $\mathbf{X}$ be $D(\mathbf{\tilde{X}})=\frac{1}{2}\sum_{v_i, v_j}\tilde{a}_{i,j}\|\mathbf{x}_i-\mathbf{x}_j\|_2^2$. Then we have
  \begin{equation}
  D(\mathbf{\tilde{X}})\leq D(\mathbf{X}).
  \end{equation}}
  Theorem 1 indicates that after aggregation, the distance between the overall connected nodes will converge in adjacent areas. While, at the same time, the unconnected nodes (in the same connected component) will also shrink together due to their common connected nodes. We define the similarities of the sum of any two nodes within a cannot-link node pair as ``negative smoothness'' and denote the total similarities of any two nodes within a must-link node pair as ``positive smoothness''. Obviously, positive smoothness brings profits while the negative smoothness is detrimental for the final node embeddings. According to the Theorem 1, we have the following Corollary 1 in dynamic graph embedding setting.\\


\noindent
\textbf{Corollary 1}. \textit{In dynamic graph setting, we use $\mathbf{\tilde{H}}_t$ and $\mathbf{H}_t$ to denote two cases: applying graph convolution to $\mathbf{X}_t$; not applying graph convolution to $\mathbf{X}_t$, respectively. Similarly, let the overall distance of hidden state at time step $t$ be $D(\mathbf{\tilde{H}}_t)=\frac{1}{2}\sum_{v_t^i, v_t^j}\tilde{a_t}_{i,j}\|\mathbf{h}_t^i-\mathbf{h}_t^j\|_2^2$. Then we have:
  \begin{equation}
  D(\mathbf{\tilde{H}}_t) \leq D(\mathbf{H}_t),
  \end{equation} 
  and
  \begin{equation}
  \sum_t D(\mathbf{\tilde{H}}_t) \leq \sum_t D(\mathbf{H}_t).
  \end{equation} 
\textbf{Proof}.
} See Appendix B. \\
The above corollary reveals that using graph convolutions in dynamic graph embedding will make the embedding space shrink in comparison with no graph convolution operation, and the shrink will increase along with time steps.

\subsection{Proposed Method}
Here, we follow the basic assumptions for dynamic graph embedding in \cite{tgcn:scalable} that (1) nodes change their hidden representations gradually over time; (2) the representation of two nodes are more similar to each other when they interact frequently than two faraway nodes, thus we have the following optimization target:

\begin{equation}
\label{TGCN:opt_target}
\begin{aligned}
\min\sum_{t=1}^{T} \sum_{i\in \mathcal{V}_t}\left \|\mathbf{z}_t^i-\mathbf{z}_{t-1}^i\right\|_2^2
&+ \sum_{(i,j)\in E}\|\mathbf{z}_t^i- \mathbf{z}_t^j\|_2^2\\
&-\lambda\sum_{(i,j)\notin E}\|\mathbf{z}_t^i- \mathbf{z}_t^j\|_2^2.
\end{aligned}
\end{equation} 

For the first term, we actually do not need to add external regularizer if we adopt temporal-smoothness or recurrent learning fashion since they have already played the role of temporal regularization; For the second term, in view of the fact that we perform at least one aggregation operation at each time step, and then the distance of connected node pairs will reduce. In other words, the second term will decrease in the condition of using GCN. Then, according to Corollary 1, the overall distance of unconnected node pairs will also decrease at the same time. Therefore, the key point is to add a constraint to the third term (e.g., increase the distance between nodes in unconnected node pairs at time step $t$ or keep it unchanged). Here, we adopt the method that is adding L2 feature normalization to push nodes on the unit hypersphere, maintaining the overall distance of $D(\mathbf{Z}_t)$ unchanged and also building connections with cosine similarity, which can be formulated as:
\begin{equation}
\label{TGCN:total distance}
\begin{aligned}
D(\mathbf{Z}_t)=\sum_{(i,j)\in E}\|\mathbf{z}_t^i- \mathbf{z}_t^j\|_2^2+\sum_{(i,j)\notin E}\|\mathbf{z}_t^i- \mathbf{z}_t^j\|_2^2 .
\end{aligned}
\end{equation} 

As mentioned above, the value of the first term in Equation (10) is reduced after aggregation in graph convolution and thus the second term in Equation (10) (or third term without negative sign in Equation (9)) would be increased on condition that we keep the overall distance $D(\mathbf{Z}_t)$ unchanged.

\begin{figure}[!tb]
  \includegraphics[width=\linewidth]{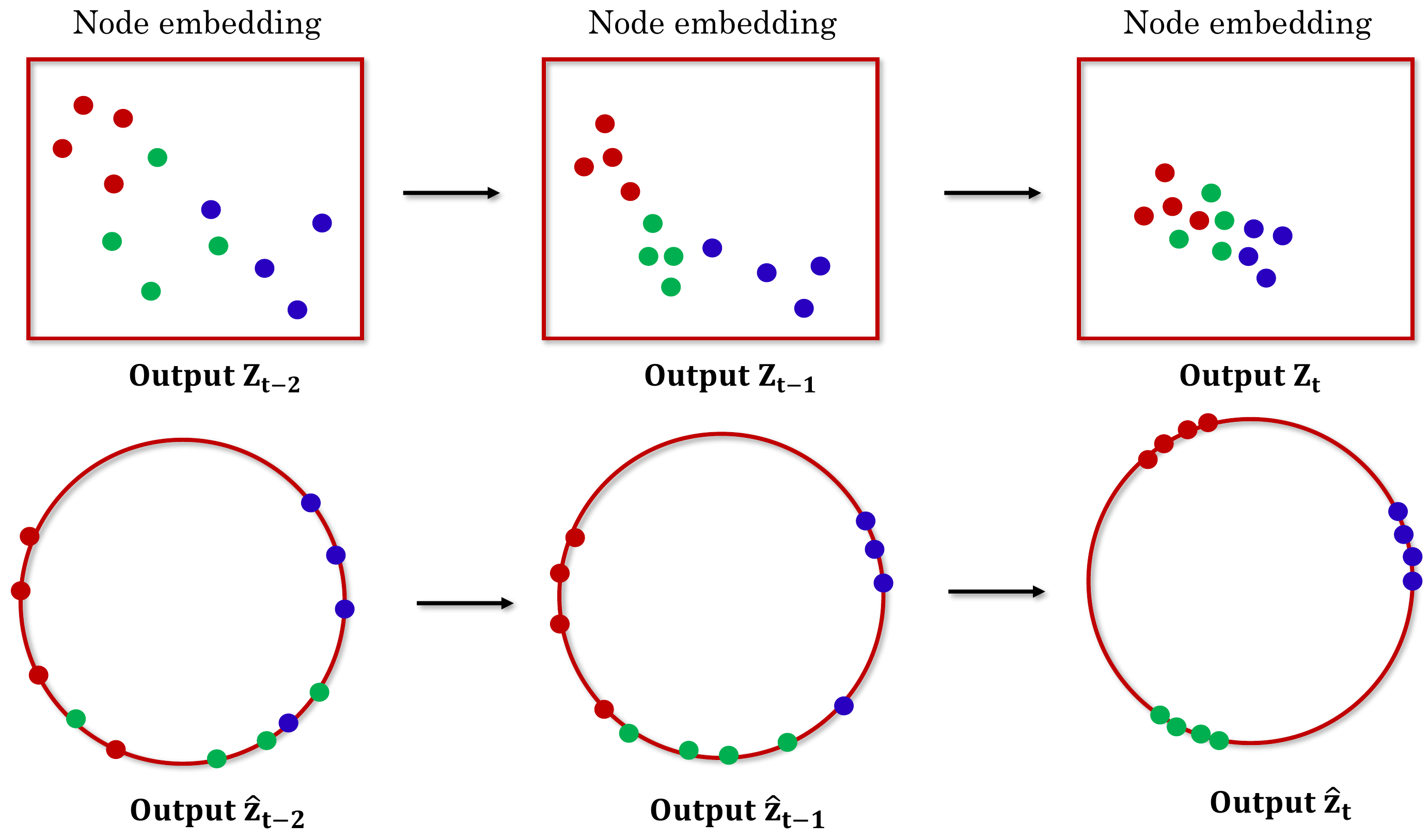}
  \caption{Illustration of the main idea for L2 feature normalization. Each sub-figure illustrates the node embeddings of the corresponding time step. The upper three sub-figures demonstrate that the unconstrained node embeddings, including all nodes, are getting closer and closer along with time steps. The lower three sub-figures show that the node embeddings with the proposed normalization constraint makes connected (or similar) nodes gather in the adjacent area while pushes unconnected (or dissimilar) nodes far away.}
  \label{fig:main_idea}
\end{figure}

Then, to keep the overall distance of node embedding unchanged, we propose to utilize L2 feature normalization, constraining the node embeddings to a unit sphere. The detailed process of our algorithm is as follows: (1) deploy a proper number (generally two or three) of graph convolutional layers, getting the raw node embeddings; (2) center the node embeddings (the reason is described in the following part); (3) place L2 normalization onto the embedding matrix so that it can be pushed to a high dimensional unit sphere, which is illustrated in Figure~\ref{fig:main_idea}; (4) feed the normalized node embeddings to next time step. Empirically, the centering step has little effect on the performance. It is worth mentioning that either using the node embeddings as the historical hidden state (recurrent learning), or as the initial value of the next input (temporal-smoothness learning), we could always achieve the above target. In this way, the must-link or similar nodes are still embedded in adjacent areas while the cannot-link or dissimilar nodes would be pushed apart.

Except for preventing all features shrinking together and being too smooth, more importantly, using L2 feature normalization can establish a direct connection between node similarity and Euclidean distance as well. By L2 feature normalization, we have the following derivation: 
\begin{equation}
\begin{aligned}
\left \|\mathbf{z}_t^i-\mathbf{z}_t^j\right\|_2 &=\sqrt{\|\mathbf{z}_t^i\|_2^2+\|\mathbf{z}_t^j\|_2^2-2(\mathbf{z}_t^i)^T\mathbf{z}_t^j)}\\
&=\sqrt{2- 2<\mathbf{z}_t^i, \mathbf{z}_t^j>},
\end{aligned}
\end{equation}
where the $<\mathbf{z}_t^i, \mathbf{z}_t^j>$ is the inner product of $\mathbf{z}_t^i$ and $\mathbf{z}_t^j$, which denotes the similarity of node $i$ and $j$, and $\mathbf{z}_t^i$ is normalized node representation, $\mathbf{z}_t^i \leftarrow \frac{\mathbf{z}_t^i}{\|\mathbf{z}_t^i\|_2}$. The above derivation demonstrates two points with large similarity are embedded close to each other while two points associated with a small similarity are far away from each other. In the following, we will illustrate details about how L2 feature normalization with centering keeps the distance of $n$ nodes in $D(\mathbf{Z}_t)$ unchanged.
\begin{equation}
\label{TGCN: opt_target}
\begin{aligned}
D(\mathbf{Z}_t)&=\sum_{(i,j)\in V}\|\mathbf{z}_t^i- \mathbf{z}_t^j\|_2^2=
\sum_{(i,j)\in V} \left(1+1-2(\mathbf{z}_t^i)^T\mathbf{z}_t^j\right)\\
&=2n^2-2\mathbf{1}^T\mathbf{Z}_t\mathbf{Z}_t^T\mathbf{1}=2n^2-2\|\mathbf{1}^T\mathbf{Z}_t\|_2^2\\
&=2n^2-\|\frac{1}{n}\sum_{i=1}^{n}\mathbf{z}_t^i\|_2^2.
\end{aligned}
\end{equation} 
Then, by centering $\mathbf{z}_t^i$,
\begin{equation}
\mathbf{z}_t^i \leftarrow \mathbf{z}_t^i-\frac{1}{n}\sum_{i=1}^{n}{\mathbf{z}_t^i}, 
\end{equation}
we can eliminate the second term in Equation (12), which is trivial since shifting would not change the distribution, so the final total distance is $2n^2$. Though $n$ is non-decreasing along time steps, the total distance within $n$ nodes is consistently $2n^2$ by the derivation of Equation (12), which means that new increasing nodes will not change the original $n$ nodes' distance.
To summary, the proposed method has two steps: (1) to center all normalized node features; and (2) to normalize each node feature in the last layer of each time step.  

\subsection{Computational Complexity}
The proposed normalization contains two parts: (1) node centralization, which take $O(|\mathcal{V}_t|)$ at each time step; (2) L2 feature normalization, which takes $O(d|\mathcal{V}_t|)$ at each time step. Therefore, the total complexity is $O(Tn)$ (assuming the number of new appeared nodes $m\ll n$).

\subsection{Basic Model and Loss Function}
\begin{figure}[!tp]
  \includegraphics[width=\linewidth]{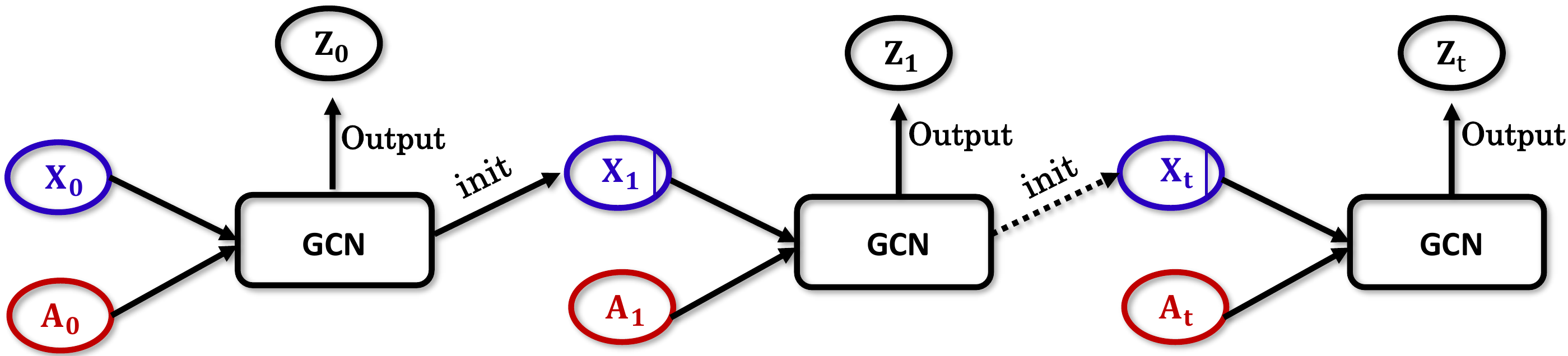}
  \caption{Illustration of temporal-smoothness learning, one typical temporal-evolving learning fashion, where the previous node hidden state is used to initialize the next-step state.}
  \label{fig:temporal-smoothness_learning}
\end{figure}

\begin{figure}[!tp]
  \includegraphics[width=\linewidth]{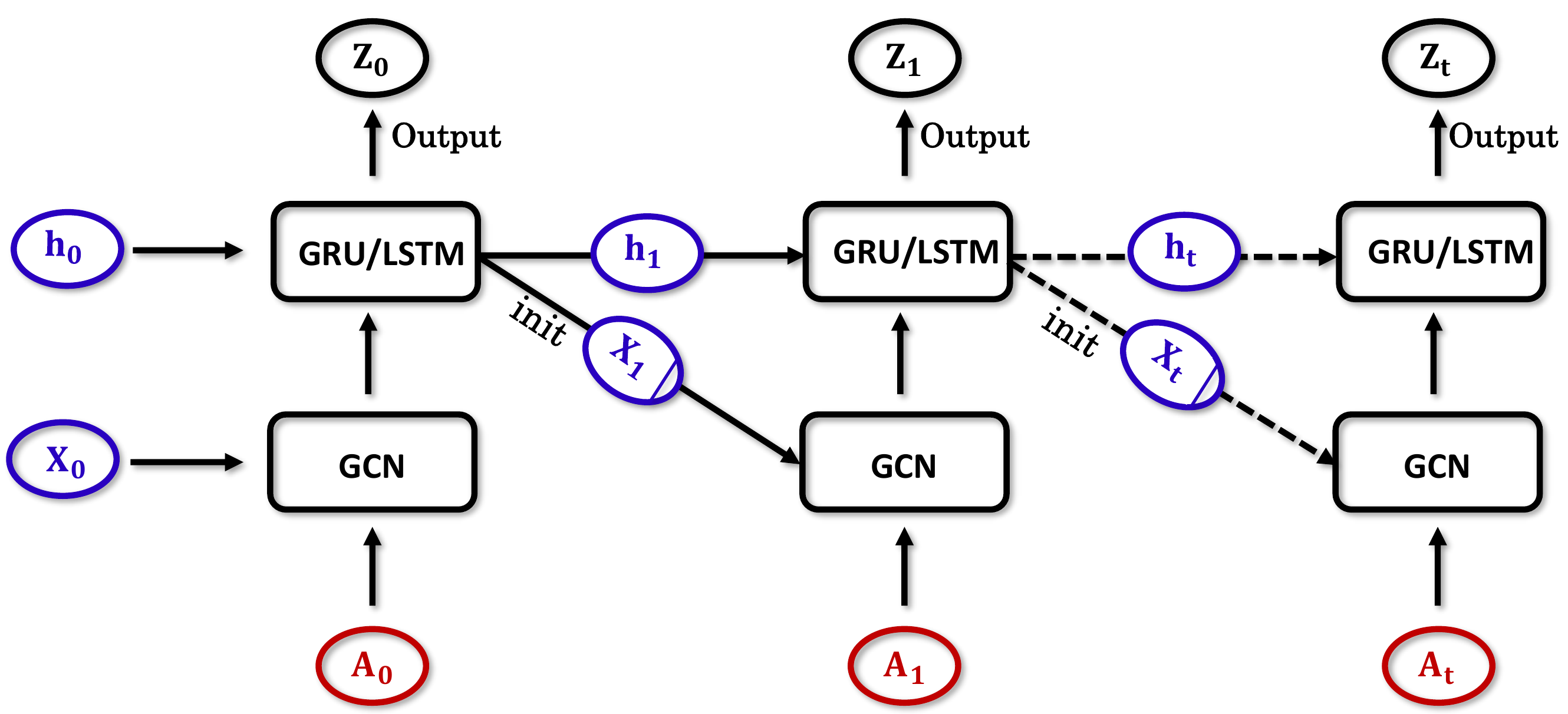}
  \caption{Illustration of recurrent learning, the other typical temporal-evolving learning fashion, where additional recurrent modules are added to capture long-distance dependencies, and the previous node hidden state is used as next node initial value.}
  \label{fig:recurrent_learning}
\end{figure}

As mentioned in \ref{section:definition of smoothness}, there are two typical ways to learn the time-evolving patterns in dynamic graph embedding. One is temporal-smoothness learning and the other is recurrent learning. In general, the former method is stable and robust while the latter is more powerful to capture long-distance dependencies. Certainly, two methods can be integrated as proposed in~\cite{tgcn:goyal2020dyngraph2vec}. 
Following the works in~\cite{tgcn:dyngem},~\cite{tgcn:gated_gcn},~\cite{tgcn:addgcn}, we extract two prototypes to verify the effectiveness of our proposed method as shown in Figure~\ref{fig:temporal-smoothness_learning} and Figure~\ref{fig:recurrent_learning}, which is also designed to exclude the influence of other factors and independently evaluate the effectiveness of the proposed method.

For clarity, we term these two frameworks as DynGCN and GRUGCN, respectively. In each snapshot $G_t(V_t, E_t)$, the input feature matrix $\mathbf{X}_t$ contains two parts: one is nodes that have ever emerged in the history, and the other is the new appear nodes. For the first part, we follow the basic idea in~ \cite{tgcn:dyngem},~\cite{tgcn:gated_gcn},~\cite{tgcn:addgcn}, that is inheriting the historical embedding results~\cite{tgcn:dyngem} as the next step input. For the second new part, we use the golort initialization method~\cite{deep_opt:glorot} to initialize nodes value. Then each GCN block is a typical two-layer graph convolution which is parameter-sharing across time steps, with adding dropout between two graph conventional layers. In Figure~\ref{fig:recurrent_learning}, an additional recurrent block is added to enhance long-distance learning, which is the prototype of models in~\cite{tgcn:gated_gcn},~\cite{tgcn:addgcn}. These two frameworks are widely used in dynamic graph embedding, currently.

To predict the occurrence of nodes appearing in the local neighborhood around $v_i$ at $t$, we use a binary cross-entropy loss function to push nodes co-occurring to have similar representation and restrain nodes not-occurring to have different representation. The loss function is:
\begin{equation} 
\label{TGCN: loss_fun}
\begin{aligned}
l_t 
&= \sum_{t}\left(\sum_{(v_i,v_j)\in E_t}-\mathrm{log}\left(\sigma({<\mathbf{z}_t^i, \mathbf{z}_t^j>})\right)\right.\\ 
&-\left.\lambda\cdot\sum_{(v_i', v_j')\notin E_t}\mathrm{log}\left(1-\sigma(<{\mathbf{z}_t^i}', {\mathbf{z}_t^j}'>)\right)\right),
\end{aligned}
\end{equation} 
where $\sigma$ is the sigmoid function and $\{(v_i', v_j')\notin E_t\}$ is a negative set sampled with negative sample strategy, and $\lambda$ is a hyperparameter. Here, we only consider negative samples that have never appeared, that is, links that appeared before time $t$ are assumed positive links, so we avoid sampling such node pairs.


\section{Experiments}
\label{section:experiment}
\subsection{Dataset}
There are numerous dynamic networks in the real world, we choose four real-world datasets that closely represent the dynamic graphs in most cases, which are publicly available: DNC\footnote{http://konect.cc/networks/dnc-temporalGraph/} and Enron\footnote{http://www.cs.cmu.edu/~enron/} email networks, UCI\footnote{http://konect.cc/networks/opsahl-ucsocial} community networks and Facebook wall posts\footnote{http://konect.cc/networks/facebook-wosn-wall}(FBW) communication networks.

\textbf{DNC} and \textbf{Enron} are two email communication networks that come from the Enron company and the 2016 Democratic National Committee email leak event, respectively.
In DNC, a node corresponds to a person and the link denotes that a person has sent an email to another person. In Enron, a node is the employee, and links are emails connections between employees. \textbf{UCI} contains sent messages between the users of an online community of students from the University of California, Irvine. A node represents a user, and the edge denotes a sent message. Multiple edges indicate multiple connections. \textbf{FBW} is a subset of posts from some users to other walls on Facebook. The nodes are the Facebook users, and an edge represents one post, linking the users writing a post to the users whose wall is written on. There also exist multiple edges pointing to a single user. Moreover, one user can be written on their walls, so we keep these links. 

There are different time spans of the four datasets, we slice each dataset into different time steps according to a roughly same number of links along the timeline. For the reason that keeping the links appeared on the same day in a time interval, the number of nodes in each time step is slightly different. We summary the dataset and the sliced time steps in TABLE \ref{table:data_summary}.
 
\begin{table}[!tp]
  \centering
  \caption{Data Summary and Sliced Time Steps}
  \resizebox{0.95\linewidth}{14.5mm}{
    \begin{tabular}{lcccc}
          \toprule
            Dataset            & DNC        & UCI        & FBW         & ENRON       \\
            \midrule
            \#Nodes           & 1,891       & 1,899       & 46,952       & 87,273       \\
            \#Links            & 39,264      & 59,835      & 876,993      & 1,148,072     \\
            \#Sliced edges      & $\sim$2,000 & $\sim$3,000 & $\sim$50,000 & $\sim$50,000 \\
            \#Total time steps & 12         & 12         & 18          & 23          \\
            \#Train steps      & 1-9   & 1-9   & 1-14   & 1-17        \\
            \#Test steps       & 10-12 & 10-12 & 15-18  & 18-22 \\ 
            \bottomrule
    \end{tabular}}
    \label{table:data_summary}
\end{table}

\begin{table*}[htpb]
    \centering
    \caption{Averaged AUC scores and AP scores across all test snapshots}
    \label{ table:all_scores}
    \resizebox{0.9\textwidth}{!}{%
        \begin{threeparttable}
        \begin{tabular}{@{}lllllllll@{}}
\toprule
& \multicolumn{2}{c}{\textbf{DNC}}                                   & \multicolumn{2}{c}{\textbf{UCI}}                                   & \multicolumn{2}{c}{\textbf{FBW}}                                   & \multicolumn{2}{c}{\textbf{Enron}}                                 \\ \cmidrule(l){2-9} 
& \multicolumn{1}{c}{\textbf{AUC}} & \multicolumn{1}{c}{\textbf{AP}} & \multicolumn{1}{c}{\textbf{AUC}} & \multicolumn{1}{c}{\textbf{AP}} & \multicolumn{1}{c}{\textbf{AUC}} & \multicolumn{1}{c}{\textbf{AP}} & \multicolumn{1}{c}{\textbf{AUC}} & \multicolumn{1}{c}{\textbf{AP}} \\ \midrule
GCN                & $91.79\pm0.47$                       & $92.20\pm0.45$                      & $73.01\pm0.66$                       & $67.06\pm0.53$                      & $71.92\pm0.40$                       & $69.52\pm0.69$                      & $88.19\pm0.29$                       & $90.79\pm0.17$                      \\
GAT                & $89.48\pm0.82$                       & $91.93\pm0.75$                      & $74.86\pm1.62$                       & $75.35\pm1.70$                      & $75.31\pm0.47$                       & $76.86\pm0.66$                      & $86.81\pm1.04$                       & $89.41\pm0.63$                      \\
SAGE               & $86.10\pm2.61$                       & $85.90\pm2.61$                      & $69.15\pm1.34$                       & $65.63\pm1.34$                      & $58.27\pm0.55$                      & $53.54\pm0.55$                      & $69.12\pm1.18$                       & $61.60\pm1.18$                      \\ \midrule

DynGCN             & $86.30\pm1.47$                       & $86.98\pm1.28$                      & $73.01\pm1.83$                       & $67.48\pm2.46$                      & $70.14\pm1.06$                       & $70.84\pm1.34$                      & $82.38\pm1.02$                       & $85.38\pm0.80$                      \\
\textbf{DynGCN+FN} & $\textbf{91.42}\pm0.30$              & $\textbf{93.39}\pm0.44$             &$\textbf{76.42}\pm0.58$              & $\textbf{74.08}\pm0.83$             &$\textbf{79.89}\pm0.71$              & $\textbf{81.25}\pm0.44$             &$\textbf{85.85}\pm0.61$              & $\textbf{89.09}\pm0.39$             \\ \midrule
GRUGCN             & $\textbf{94.12}\pm1.94$                       & $\textbf{95.11}\pm0.95$             & $72.57\pm1.47$                       & $68.57\pm4.20$                      & $71.41\pm2.40$                       & $73.25\pm2.35$                      & $84.17\pm2.87$                       & $88.60\pm1.38$                      \\
$\textbf{GRUGCN+FN}$ & $93.02\pm0.62$              & $94.86\pm0.42$             & $\textbf{75.41}\pm2.10$              & $\textbf{74.87}\pm1.89$             & $\textbf{79.44}\pm0.39$              & $\textbf{82.58}\pm0.35$             & $\textbf{87.88}\pm0.44$              & $\textbf{90.75}\pm0.35$             \\ \midrule
EGCO    & $87.43\pm0.60$                       & $88.33\pm0.63$                      &  $\textbf{74.16}\pm3.78$                       & $70.69\pm5.10$                      &  $66.46\pm0.36$                       & $63.71\pm0.77$                      & $81.62\pm0.32$                       & $79.68\pm0.44$                      \\
\textbf{EGCO+FN}   & $\textbf{92.65}\pm0.64$              & $\textbf{94.74}\pm0.42$             & $73.70\pm2.41$              & $\textbf{75.25}\pm1.80$             & $\textbf{72.30}\pm0.38$              & $\textbf{76.09}\pm0.57$             & $\textbf{87.72}\pm0.54$              & $\textbf{90.95}\pm0.35$             \\ \midrule
EGCH    & $89.27\pm0.41$                       & $90.72\pm0.49$                      & $\textbf{78.81}\pm1.95$                       & $\textbf{77.16}\pm0.83$          & $69.65\pm0.31$                       & $69.48\pm0.49$                      & $82.06\pm0.21$                       & $80.03\pm0.14$                      \\
\textbf{EGCH+FN}   & $\textbf{92.78}\pm0.31$              & $\textbf{95.08}\pm0.13$             & $74.65\pm2.91$              & $76.41\pm1.73$             & $\textbf{74.34}\pm0.44$              & $\textbf{78.64}\pm0.21$             & $\textbf{87.72}\pm0.51$              & $\textbf{91.35}\pm0.32$             \\ \bottomrule
\end{tabular}%
\begin{tablenotes}
 \item[*] The best results of the proposed L2 feature normalization (+FN) and basic dynamic graph models are bold.
\end{tablenotes}
\end{threeparttable}
}
\end{table*}

\subsection{Comparison Models}
Here we compare a variety of competing unsupervised node embedding methods. 
First, we compare the typical static graph embedding models: GCN~\cite{gcn:vae} which consists of two-layer graph convolution, using an encoder-decoder framework to learn node embeddings; GAT~\cite{gcn:gat} is a variation of GCN where each graph convolution is replaced with attention mechanism; SAGE~\cite{gcn:graphsage} is an inductive framework that combine node attribute information to learn representations based on previously unseen graph data. Then, we utilize the basic frameworks DynGCN and GRUGCN mentioned above, to verify our proposed normalization method. Furthermore, we compare against the latest released work EvolveGCN~\cite{tgcn:evolvegcn}, which has two versions EGCNO and EGCNH. EvolveGCN uses RNN to evolve GCN parameters, aiming at preserving temporal patterns in its weights.


\subsection{Evaluation Tasks and Metric}
In dynamic graph embedding, link prediction is one of the most common methods used to evaluate the quality of node embedding. Dynamic link prediction is defined as given $G=\{G_1, \cdots, G_{t}\}$, to predict links in the future $m$ snapshots $\{G_{t + 1}, G_{t+2},\cdots, G_{t+m}\}$, including ever appeared links (transductive learning) and new links (inductive learning). The links that appeared in $\{G_{t + 1}, G_{t+2},\cdots, G_{t+m}\}$ are considered as positive samples and node pairs without links are as negative samples. Then the models are evaluated by correctly classifying positive and negative links. We sample an equal amount of negative node pairs and compute the average precision (termed as AP) and area under the ROC curve (termed as AUC) scores as the metric. 

In addition, we define the ratio of negative-to-positive smoothness as an auxiliary metric to evaluate the model. The target of node embedding is to make similar nodes cluster together and embed in the adjacent area while push nodes of no similarity are far away from each other. Therefore, the ratio of negative-to-positive smoothness can well evaluate the final embeddings. In addition, the ratio of negative-to-positive smoothness shows how much the negative effects of Laplacian smoothing in graph convolution-based models brings, which is defined as:
\begin{equation}
R_{neg} =\frac{n_s}{p_s},\\
\end{equation} 
where $p_s=\sum_{(i,j) \in {E}_t}\textbf{z}_t^i\textbf{z}_t^j$ and $
n_s=\sum_{(i,j)\notin {E}_t}{\textbf{z}_t^i\textbf{z}_t^j}$.
Since the number of unconnected nodes is huge, we randomly sample an equal number of negative samples and average their similarities. 

\subsection{Experimental Setup}
We roughly split 80\% graph snapshots for training and the rest 20\% snapshots for testing, and the detail is shown in Table~\ref{table:data_summary}. If no specification, we set the same experimental setting for each model. During training, we utilize Adam SGD optimizer~\cite{Adam} with weight decay and the $\lambda$ value is $ 5\times 10^{-7}$ to optimize the loss function. At each time step, we add dropout~\cite{opt:dropout} before each graph convolutional layer, and the dropout rate is 0.25. 
To fully utilize the time information, we do not split any validation set to tune the parameters. Instead, we select the model based on the loss value during training. The training epoch is fixed as 100. For new appear nodes in the time step $t$, the initial value is initialized with dimension $n\times 32$ ($n=|\mathcal{V}_t|$) if there is no feature available. The dimension of the middle and the output layer is set to 32, which is to reduce the effort of hyperparameter tuning. Finally, we randomly sample 5 different seeds and repeat the experiment. Our code is publicly available \url{https://github.com/marlin-codes/FeatureNorm}


\subsection{Experimental Results}
Table~\ref{ table:all_scores} summarizes the averaged AUC scores and AP scores across all test snapshots on four datasets. From the results, we have the following observations: (1) Being equipped with the proposed normalization method, the baseline models, including DynGCN, GRUGCN, EGCNO, and EGCNH, are all improved obviously in most cases; (2) In comparison with the static graph embedding, the dynamic graph models obtain higher AUC and AP scores, indicating the informativeness of temporal-evolving patterns; (3) Some static graph models outperform dynamic embedding in several non-constraint cases (e.g., AUC scores are from GCN and DynGCN performing on DNC/UCI/FBW/ENRON datasets) while being defeated or being improved much by our proposed method (+FN), demonstrating that the damage of incremental feature shrink and the constructiveness of our proposed normalization; (4) Being deployed with L2 feature normalization, taking the performance of DynGCN on the four datasets as an example, the proposed models achieve 3\%$\sim$10\% improvements; (5) The L2 normalization can also benefit robustness, where the evidence can be found by comparing the standard variance of baseline model with the normalized version, which further demonstrates the superiority of the proposed normalization. 
\begin{table*}[htpb]
\centering
\caption{The ratio of negative smoothness-to-positive smoothness}
\label{tab:my-table}
\resizebox{0.95\textwidth}{!}{
\begin{threeparttable}
\begin{tabular}{@{}lcccccccc@{}}
\toprule
\multicolumn{1}{l}{\multirow{2}{*}{}} & \multicolumn{2}{c}{DNC} & \multicolumn{2}{c}{UCI} & \multicolumn{2}{c}{FBW} & \multicolumn{2}{c}{Enron} \\
\multicolumn{1}{l}{} & Average $R_{neg}$ & Latest $R_{neg}$ & Average $R_{neg}$ & Latest $R_{neg}$ & Average $R_{neg}$ & Latest $R_{neg}$ & Average $R_{neg}$ & Latest $R_{neg}$ \\ \midrule
DynGCN & $0.4591$ & $0.4667$ & $0.2265$ & $0.1601$ & $0.1527$ & $0.1018$ & $0.4198$ & $0.1820$ \\
\textbf{DynGCN+FN} & $\textbf{0.0277}$ & $\textbf{0.0123}$ & $\textbf{0.0766}$ & $\textbf{0.0115}$ & $\textbf{0.0069}$ & $\textbf{0.0025}$ & $\textbf{0.0346}$ & $\textbf{0.0291}$ \\ \midrule
GRUGCN & $0.9444$ & $0.9255$ & $0.1157$ & $0.0685$ & $0.0935$ & $0.1050$ & $0.2026$ & $0.3844$ \\
\textbf{GRUGCN+FN} & $\textbf{0.0138}$ & $\textbf{0.0246}$ & $\textbf{0.0296}$ & $\textbf{0.0094}$ & $\textbf{0.0031}$ & $\textbf{0.0029}$ & $\textbf{0.0397}$ & $\textbf{0.0119}$ \\ \midrule
EGCO & $0.0183$ & $0.0146$ & $\textbf{0.0150}$ & $0.0162$ & $0.0388$ & $0.0132$ & $0.1454$ & $0.083$ \\
\textbf{EGCO+FN} & $\textbf{0.0058}$ & $\textbf{0.0006}$ & $0.0181$ & $\textbf{0.0143}$ & $\textbf{0.0019}$ & $\textbf{0.0010}$ & $\textbf{0.0098}$ & $\textbf{0.0081}$ \\ \midrule
EGCH & $0.0329$ & $0.0270$ & $0.0139$ & $0.0198$ & $0.0344$ & $0.0254$ & $0.1685$ & $0.1361$ \\
\textbf{EGCH+FN} & $\textbf{0.0064}$ & $\textbf{0.0070}$ & $\textbf{0.0153}$ & $\textbf{0.0090}$ & $\textbf{0.0016}$ & $\textbf{0.0007}$ & $\textbf{0.0088}$ & $\textbf{0.0047}$ \\ \bottomrule
\end{tabular}%
\begin{tablenotes}
 \item[*] The best results of the proposed L2 feature normalization (+FN) and basic dynamic graph models are bold.
\end{tablenotes}
\end{threeparttable}
\label{tab: smoothness}
}
\end{table*}

Table~\ref{tab: smoothness} shows the ratio of negative smoothness to positive smoothness. The results contain two parts, one is the averaged ratio across all training time spans, which reflects the whole node embedding quality. The other part is the final embedding representation which is the latest one used for evaluation. From the results shown in Table~\ref{tab: smoothness}, we discover that after applying L2 feature normalization, the ratio of negative smoothness to positive smoothness is dramatically reduced in most cases. Certainly, the decreasing negative smoothness means that the ratio of positive smoothness is increasing. Yet, we also find in some cases, taking performance of ECGNH on UCI dataset as an example where the ratio is not reduced after the normalization but the interesting thing is the corresponding performance (in Table~\ref{ table:all_scores}) is also not improved. Thus, we further confirm that feature shrink is part of the reason for relatively low performance while our proposed method can unleash models' true capacity.

\begin{figure*}[t!]
    \centering
    \includegraphics[width=0.95\linewidth]{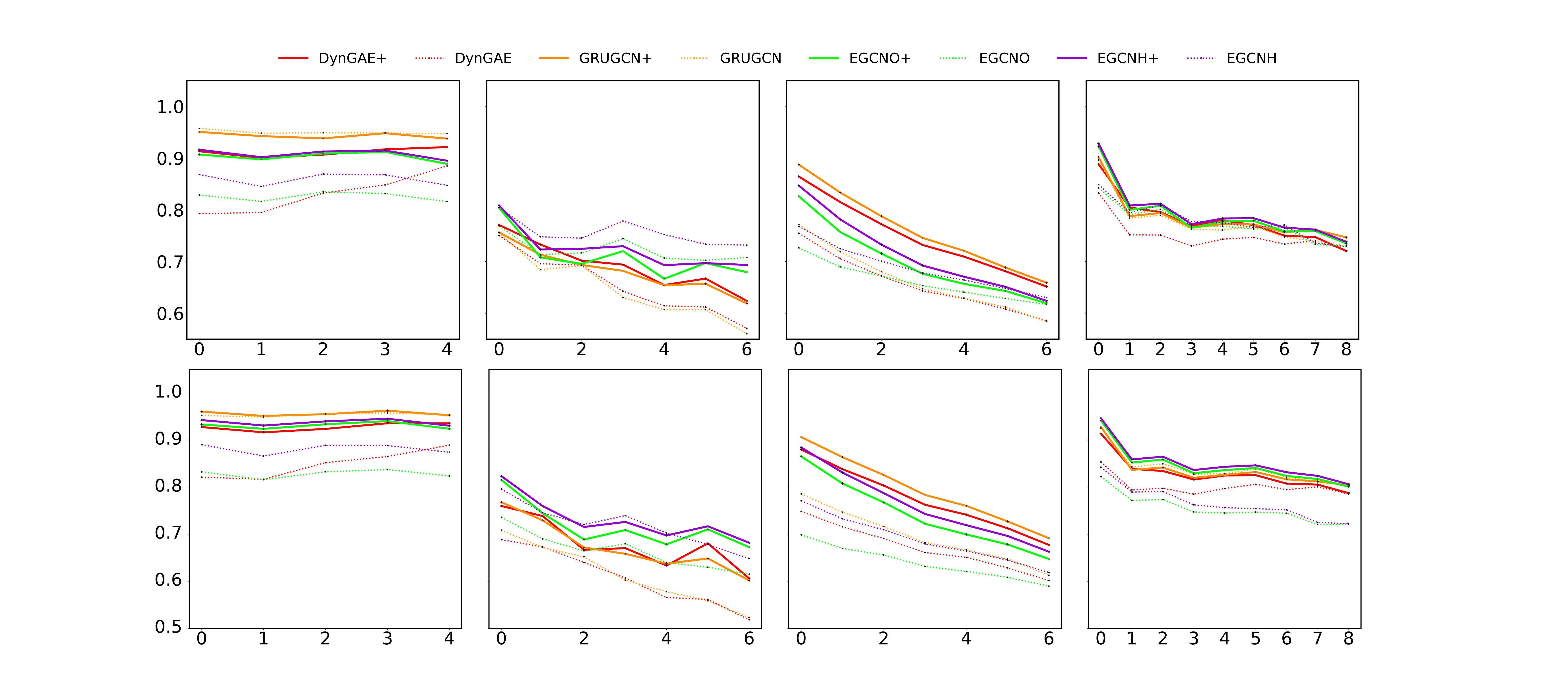}
    \caption{AUC (the upper four sub-figures) and AP (the lower four sub-figures) scores for long time-step prediction. From left to right, the corresponding dataset is DNC, UCI, FBW and Enron.}
    \label{fig:aucap}
\end{figure*}

\section{Analysis and Discussion}
Although some parameters like learning rate, dropout rate, embedding dimension can affect the final performance, they actually show consistent performance with the same environmental setting which is not the main concern in our paper. Here, we care more about the performance of long-distance prediction and the difference with the related normalization method. 

\subsection{Long Distance Learning}
Long-distance prediction is essential for some practical scenarios, which can also indicate the embedding qualities include stability and robustness. In the last section, we used 80\% and 20\% of the dataset for training and test, respectively. Here we further extend the steps of the test data, splitting 60\% and 40\% for model learning and evaluation, respectively. The results are shown in Figure~\ref{fig:aucap} including AUC scores and AP scores. Based on the results, we have the following findings: (1) There is a descending tendency in the performance of AUC and AP along with time steps, which indicates the dependencies of time-evolving patterns matters; (2) Adding the proposed normalization benefits the embeddings from two aspects: one is to obtain a better or comparable performance and the other is long durability since models with the proposed method are able to obtain high AUC and AP scores as well as time goes by.
\begin{table*}[htpb]
    \centering
    \caption{Averaged AUC scores and AP scores with different normalization methods}
    \label{table:pn_fn_no}
    \resizebox{0.9\textwidth}{!}{%
    \begin{threeparttable}
        \begin{tabular}{@{}lllllllll@{}}
            \toprule
            & \multicolumn{2}{c}{\textbf{DNC}}                                   & \multicolumn{2}{c}{\textbf{UCI}}                                   & \multicolumn{2}{c}{\textbf{FBW}}                                   & \multicolumn{2}{c}{\textbf{Enron}}                                 \\ \cmidrule(l){2-9} 
            & \multicolumn{1}{c}{\textbf{AUC}} & \multicolumn{1}{c}{\textbf{AP}} & \multicolumn{1}{c}{\textbf{AUC}} & \multicolumn{1}{c}{\textbf{AP}} & \multicolumn{1}{c}{\textbf{AUC}} & \multicolumn{1}{c}{\textbf{AP}} & \multicolumn{1}{c}{\textbf{AUC}} & \multicolumn{1}{c}{\textbf{AP}} \\ \midrule
            DynGCN             
            &$86.30\pm1.47$                       &$86.98\pm1.28$                      &$73.01\pm1.83$                       &$67.48\pm2.46$                      &$70.14\pm1.06$                       &$70.84\pm1.34$                 
            &$82.38\pm1.02$                       &$85.38\pm0.80$                      \\
            DynGCN+PN       
            &$82.35\pm2.04$                       &$83.63\pm1.71$                      &$68.99\pm4.35$                       &$59.07\pm4.25$                      &$63.35\pm4.22$                       &$63.87\pm3.12$                  
            &$84.66\pm2.43$                       &$86.80\pm2.32$                      \\
            DynGCN+PN-SI       
            &$85.16\pm1.27$                       &$85.17\pm1.32$                      &$69.46\pm4.00$                       &$60.17\pm3.83$                      &$62.73\pm4.37$                       &$63.12\pm3.34$                      &$83.18\pm4.50$                       &$86.16\pm3.21$  \\
            \textbf{DynGCN+FN} 
            &$\textbf{91.42}\pm0.30$              &$\textbf{93.39}\pm0.44$             &$\textbf{76.42}\pm0.58$              &$\textbf{74.08}\pm0.83$             &$\textbf{79.89}\pm0.71$              &$\textbf{81.25}\pm0.44$             &$\textbf{85.85}\pm0.61$              &$\textbf{89.09}\pm0.39$             
            \\ \midrule
            GRUGCN             
            &$\textbf{94.12}\pm1.94$                      &$\textbf{95.11}\pm0.95$                      &$72.57\pm1.47$                       &$68.57\pm4.20$                      &$71.41\pm2.40$                       &$73.25\pm2.35$                      &$84.17\pm2.87$                       &$88.60\pm1.38$                      \\
            GRUGCN+PN          
            &$91.55\pm2.10$                       &$93.45\pm1.30$                      &$74.48\pm4.18$                       &$70.22\pm4.80$                      &$73.73\pm4.84$                       &$76.99\pm4.53$                      &$84.37\pm4.07$                       &$86.85\pm3.57$                      \\
            GRUGCN+PN-SI       
            &$89.93\pm2.89$                       &$92.66\pm1.68$                      &$73.64\pm4.34$                       &$69.79\pm5.66$                      &$74.18\pm4.56$                       &$76.63\pm4.09$                      &$84.26\pm4.16$                       &$86.39\pm3.55$                      \\
            \textbf{GRUGCN+FN} 
            &$93.02\pm0.62$              
            &$94.86\pm0.42$             &$\textbf{75.41}\pm2.10$              &$\textbf{74.87}\pm1.89$            &$\textbf{79.44}\pm0.39$              &$\textbf{82.58}\pm0.35$             &$\textbf{87.88}\pm0.44$              &$\textbf{90.75}\pm0.35$             \\ \midrule
            EGCO               
            &$87.43\pm0.60$                       &$88.33\pm0.63$                      &$\textbf{74.16}\pm3.78$                       &$70.69\pm5.10$                      &$66.46\pm0.36$                       &$63.71\pm0.77$                      &$81.62\pm0.32$                       &$79.68\pm0.44$                      \\
            EGCO+PN            
            &$77.95\pm2.79$                       &$73.28\pm4.19$                      &$64.14\pm3.21$                       &$56.06\pm2.90$                      &$63.59\pm1.48$                       &$59.06\pm1.21$                      &$75.35\pm2.30$                       &$69.98\pm1.84$                      \\
            EGCO+PN-SI         
            &$78.59\pm2.79$                       &$74.32\pm4.18$                      &$65.35\pm3.31$                       &$56.68\pm2.90$                      &$63.88\pm1.41$                       &$59.25\pm1.12$                      &$75.70\pm2.39$                       &$70.43\pm2.05$                      \\
            \textbf{EGCO+FN}   
            &$\textbf{92.65}\pm0.64$              &$\textbf{94.74}\pm0.42$             &$73.70\pm2.41$              &$\textbf{75.25}\pm1.80$             &$\textbf{72.30}\pm0.38$              &$\textbf{76.09}\pm0.57$             &$\textbf{87.72}\pm0.54$              &$\textbf{90.95}\pm0.35$             \\ \midrule
            EGCH               
            &$89.27\pm0.41$                       &$90.72\pm0.49$                      &$\textbf{78.81}\pm1.95$                       &$\textbf{77.16}\pm0.83$                      &$69.65\pm0.31$                       &$69.48\pm0.49$                      &$82.06\pm0.21$                       &$80.03\pm0.14$                      \\
            EGCH+PN            
            &$83.64\pm2.28$                       &$83.35\pm3.69$                      &$68.20\pm4.47$                       &$63.28\pm3.32$                      &$65.75\pm2.25$                       &$62.75\pm2.03$                      &$80.19\pm2.43$                       &$78.20\pm2.67$                      \\
            EGCN+PN-SI         
            &$83.79\pm2.35$                       &$83.68\pm3.61$                      &$68.99\pm4.85$                       &$63.92\pm3.64$                      &$66.32\pm2.32$                       &$63.75\pm2.40$                      &$79.85\pm2.52$                       &$77.38\pm2.85$                      \\
            \textbf{EGCH+FN}   
            &$\textbf{92.78}\pm0.31$              &$\textbf{95.08}\pm0.13$             &$74.65\pm2.91$              
            &$76.41\pm1.73$             &$\textbf{74.34}\pm0.44$              &$\textbf{78.64}\pm0.21$             &$\textbf{87.72}\pm0.51$              &$\textbf{91.35}\pm0.32$            \\ \bottomrule
        \end{tabular}
 \begin{tablenotes}
 \item[*] The best results of each model with different constraints are bold.
\end{tablenotes}
\end{threeparttable}
    }
\end{table*}
\subsection{Difference with Related Works}
The underlying principle to constrain the overall distance unchanged of the proposed method is similar with PairNorm~\cite{gcn_opt:pairnorm}. So it is necessary to point out the difference and make an experimental comparison. 

From the task aspect, PairNorm is designed for deeper graph neural networks, evaluated on the node classification task while we care more for similarity for node embeddings and dynamic link predictions. The difference will result in distinct designs for the corresponding settings and tasks.
From the model design aspect, PairNorm is proposed to keep the total pairwise distance unchanged in the background of a static graph. However, in a dynamic setting, we need to consider that the number of nodes in each time can be different. 
From the implementation aspect, PairNorm adds the constraint between graph convolutional layer\footnote{https://github.com/LingxiaoShawn/PairNorm/blob/master/models.py}. Then, the following operations, including softmax and classifier, will change the total distance and similarities. However, our normalization is assigned to the final layer of graph convolutions in each time step which pushes the embeddings on the
unit hypersphere and the total distance keeps unchanged. In addition, the proposal also builds a connection between cosine similarity and embedding distance. In the end, we impose PairNorm (PN) and PairNorm-SI (PN-SI) mentioned in~\cite{gcn_opt:pairnorm} to dynamic graph models for an experimental comparison. 

According to the AUC and AP scores shown in Table~\ref{table:pn_fn_no}, we can conclude that PairNorm does not work in the task of dynamic graph embedding, and even hinders the performance of the model. However, our proposed normalization method can significantly improve the AUC and AP scores of the original model.




\section{Conclusions}
In this paper, we inspect the feature shrinking and oversmoothing problem in dynamic graph embedding. First, we analyze two basic operations in graph convolution, and then come to two typical learning frameworks in dynamic graph embedding: temporal-smoothness learning and recurrent learning.
By theoretical analysis on overall embedding distance, we discover nodes including must-link pairs and cannot-link pairs would get closer if there is no constraint applied to graph convolution. Actually, most dynamic graph embedding models with graph convolution exist such a problem, but seldom works notice it. In this work, we propose L2 feature normalization to constrain the overall distance of pairwise nodes at time step $t$ unchanged, which can be also regarded as to constrain overall similarity unchanged. In this way, the positive smoothness will be further improved while negative smoothness can be reduced. In addition, the proposal can be regarded as a plug-in module that is easy to extend related dynamic graph embedding models.
\section*{Acknowledgements}
The work described in this paper was partially supported by the Research Grants Council of the Hong Kong Special Administrative Region, China (CUHK 2410021, Research Impact Fund, No. R5034-18). We would like to thank the anonymous reviewers for their comments. Besides, we gratefully thank Xue Li for his helpful feedback and discussions.

\section*{Appendix}
For simplicity and clarity, we assume the dimension of node feature is one and it can easily extend multidimensional case since $D(\mathbf{X})/D(\mathbf{\tilde{X}})$ or $D(\mathbf{H}_t)/D(\mathbf{\tilde{H}}_t)$ can be decomposed into the sum of one-dimension case. In the following, we use the corresponding lowercase in case of confusion.

\subsection*{A. Proof of Theorem 1}
Here we rewrite the proof in line with the following corollary 1 for better understanding feature shrink in dynamic graph embedding. \\

\noindent    
\textbf{Theorem 1}~\cite{labelpropagation}
\textit{Let the overall distance of node embeddings be $D(\mathbf{\tilde{x}})=\frac{1}{2}\sum_{v_i, v_j}\tilde{a}_{i,j}\|\mathbf{x}_i-\mathbf{x}_j\|_2^2$. Then we have
\begin{equation}
D(\mathbf{\tilde{x}})\leq D(\mathbf{x}).
\end{equation}}\\
\noindent
To prove the Theorem 1, two lemmas related to gradient of $D(\mathbf{x})$ and Hessian matrix of $D(\mathbf{x})$ will be introduced at first.\\

\noindent
\textbf{Lemma 1} \textit{Gradient of $D(\mathbf{x})$ satisfies:} $ \tilde{\mathbf{x}}_{i}-\frac{\partial D\left(\mathbf{x}\right)}{\partial \mathbf{x}_{i}} = \mathbf{x}_{i}$\\
\noindent
\textit{Proof}:\\
\begin{equation}
\begin{aligned}
\quad \tilde{\mathbf{x}}_{i}-\frac{\partial D\left(\mathbf{x}\right)}{\partial \mathbf{x}_{i}}&=\mathbf{x}_{i}-\sum_{v_{j} \in \mathcal{N}\left(v_{i}\right)} \tilde{a}_{i j}\left(\mathbf{x}_{i}-\mathbf{x}_{j}\right)\\
&=\sum_{v_{j} \in \mathcal{N}\left(v_{i}\right)} \tilde{a}_{i j} \mathbf{x}_{j}\\
&=\mathbf{x}_{i}.
\end{aligned}
\end{equation}

\noindent
\textbf{Lemma 2} \textit{Hessian matrix of $D(\mathbf{x})$ satisfies : }$ \nabla^{2} D(\mathbf{x}) \preceq 2 I$\\
\textit{Proof}: Hessian matrix of $ D(\mathbf{x})=\frac{1}{2} \sum_{v_{i}, v_{j}} \tilde{a}_{i j}\left\|\mathbf{x}_{i}-\mathbf{x}_{j}\right\|_{2}^{2} $ can be written as 
\begin{equation}
\begin{aligned}
\nabla^{2} D(\mathbf{x})=\left[\begin{array}{cccc}
1-\tilde{a}_{11} & -\tilde{a}_{12} & \cdots & -\tilde{a}_{1 n} \\
-\tilde{a}_{21} & 1-\tilde{a}_{22} & \cdots & -\tilde{a}_{2 n} \\
\vdots & \vdots & \ddots & \vdots \\
-\tilde{a}_{n 1} & -\tilde{a}_{n 2} & \cdots & 1-\tilde{a}_{n n}
\end{array}\right]\\
=I-D^{-1} A,
\end{aligned}
\end{equation}
and we then can obtain $ 2 I-\nabla^{2} D(\mathbf{x})=I+D^{-1} A .$ Here, $D^{-1} A $ is transition matrix, where each entry is non-negative, the sum of each row is one and its eigenvalues are within the range $[-1,1]$, so the eigenvalues of $ I+D^{-1} A $ are within the range $[0, 2]$. That is, $I+D^{-1} A $ is a positive semidefinite matrix, and thus $\nabla^{2} D(\mathbf{x}) \preceq 2 I $ holds on.\\

\noindent
\textit{Proof of Theorem 1}. $D$ is a quadratic function, by second-order Taylor expansion of $D $ around $ \mathbf{x} $, we have the following inequality:
\begin{small}
\begin{equation}
\begin{aligned}
D\left(\tilde{\mathbf{x}}\right) &=D\left(\mathbf{x}\right)+\nabla D\left(\mathbf{x}\right)^{\top}\left(\tilde{\mathbf{x}}-\mathbf{x}\right)+\frac{1}{2}\left(\tilde{\mathbf{x}}-\mathbf{x}\right)^{\top} \nabla^{2} D(\mathbf{x})\left(\tilde{\mathbf{x}}-\mathbf{x}\right) \\
&=D\left(\mathbf{x}\right)-\nabla D\left(\mathbf{x}\right)^{\top} \nabla D\left(\mathbf{x}\right)+\frac{1}{2} \nabla D\left(\mathbf{x}\right)^{\top} \nabla^{2} D(\mathbf{x}) \nabla D\left(\mathbf{x}\right) \\
& \leq D\left(\mathbf{x}\right)-\nabla D\left(\mathbf{x}\right)^{\top} \nabla D\left(\mathbf{x}\right)+\nabla D\left(\mathbf{x}\right)^{\top} \nabla D\left(\mathbf{x}\right)=D\left(\mathbf{x}\right).
\end{aligned}
\end{equation}
\end{small}

\subsection*{B. Proof of Corollary 1}
\noindent
\textbf{Corollary 1}. \textit{In dynamic graph embedding, we use $\mathbf{\tilde{h}}_t$ and $\mathbf{h}_t$ to denote two cases: applying graph convolution to $\mathbf{x}_t$; not applying graph convolution to $\mathbf{x}_t$, respectively. Similarly, let the overall distance of pairwise nodes state at time step $t$ be $D(\mathbf{\tilde{h}}_t)=\frac{1}{2}\sum_{v_t^i, v_t^j}\tilde{a_t}\|\mathbf{h}_t^i-\mathbf{h}_t^j\|_2^2$. Then we have:
    \begin{equation}
    D(\mathbf{\tilde{h}}_t) \leq D(\mathbf{h}_t),
    \end{equation} 
    and
    \begin{equation}
    \sum_t D(\mathbf{\tilde{h}}_t) \leq \sum_t D(\mathbf{h}_t).
    \end{equation} 
}
\\ \noindent
\textit{Proof}: (1) In temporal-smoothness learning fashion,  $\mathbf{\tilde{h}}_t=\mathbf{\tilde{x}}_t=\mathrm{GCN}(\mathbf{x}_t, \mathbf{A}_t)$ if using graph convolution, and $\mathbf{h}_t=\mathbf{x}_t$ if no graph convolution, thus we can know that $D(\mathbf{\tilde{h}}_t) \leq D(\mathbf{h}_t)$ from Theorem 1, then $\sum_t(D(\mathbf{\tilde{h}}_t) \leq D(\mathbf{h}_t))$. (2) In recurrent learning manner, we use the vanilla RNN, ignoring bias and activation, to show it. Here, $\mathbf{\tilde{h}}_t =W\mathbf{\tilde{x}}_t+U\mathbf{h}_{t-1}$ (using graph convolution) and $\mathbf{h}_t =W\mathbf{x}_t+U\mathbf{h}_{t-1}$ (no graph convolution), then we have $\mathbf{h}_t- \mathbf{\tilde{h}}_t = W(\mathbf{x}_t - \mathbf{\tilde{x}}_t)$. We can attribute the difference of $\mathbf{h}_t$ and  $\mathbf{\tilde{h}}_t$ to $\mathbf{x}_t$ and $\mathbf{\tilde{x}}_t$, where W is a transformation vector/matrix, which can be incorporated in transformation step in graph convolution. In Theorem 1, we know $\mathbf{\tilde{x}}_t$ is shrunk from $\mathbf{x}_t$, therefore we can easy to know that $\mathbf{\tilde{h}}_t$ will be shrunk from $\mathbf{h}_t$. With time increasing, the shrink will be amplified gradually if we initialize $\mathbf{x}_t$ from the last time output $\mathbf{h}_{t-1}$, and the shrink appears in each time step independently which is still accumulated if we keep using the original feature of $\mathbf{x}_t$ at each time step. Together, we have $D(\mathbf{\tilde{h}}_t) \leq D(\mathbf{h}_t)$ and $\sum_t(D(\mathbf{\tilde{h}}_t) \leq D(\mathbf{h}_t))$.




%
\bibliographystyle{IEEEtran}
\bibliography{references}
\end{document}